\pdfoutput=1

\documentclass[10pt]{revtex4-1}
\usepackage{caption}
\usepackage{subcaption}
\usepackage{mathrsfs}
\usepackage{nicefrac}
\usepackage{amssymb}
\usepackage{comment}

\usepackage{graphicx,bm,amsmath}
\usepackage[usenames,dvipsnames]{color}
\usepackage[colorlinks,bookmarks=false,citecolor=blue,linkcolor=red,urlcolor=blue]{hyperref}
\usepackage[bbgreekl]{mathbbol}
\usepackage{xcolor}

\usepackage{fancyhdr}
\usepackage{hyperref}
\usepackage[normalem]{ulem}

\usepackage{tikz}
\usepackage{listings}

\usetikzlibrary{arrows}
\usetikzlibrary{math}
\usetikzlibrary{calc}
\usetikzlibrary{shapes.geometric}

\usepackage{bbold}

\newcommand{\be}{\begin{equation}}
\newcommand{\ee}{\end{equation}}
\newcommand{\bea}{\begin{eqnarray}}
\newcommand{\eea}{\end{eqnarray}}

\definecolor{smoothred}{HTML}{C5232F}
\definecolor{mygreen}{rgb}{0,0.5,0}
\definecolor{myblue}{rgb}{0,0,0.75}
\definecolor{mymagenta}{cmyk}{0,1,0,0.12}

\def\doi{http://dx.doi.org/}

\begin{document}

\title{Quantum-inspired Machine Learning on high-energy physics data}

\author{Timo Felser}
\affiliation{Tensor Solutions, Institute for Complex Quantum Systems, University of Ulm, D-89069 Ulm, Germany}
\affiliation{Dipartimento di Fisica e Astronomia ``G. Galilei'', Universit\`a di Padova, I-35131 Padova, Italy}
\affiliation{INFN, Sezione di Padova, I-35131 Padova, Italy} 
\affiliation{Theoretische Physik, Universit\"at des Saarlandes, D-66123 Saarbr\"ucken, Germany.}

\author{Marco Trenti}
\affiliation{Tensor Solutions, Institute for Complex Quantum Systems, University of Ulm, D-89069 Ulm, Germany}
\affiliation{Dipartimento di Fisica e Astronomia ``G. Galilei'', Universit\`a di Padova, I-35131 Padova, Italy.}

\author{Lorenzo Sestini}
\affiliation{INFN, Sezione di Padova, I-35131 Padova, Italy.}

\author{Alessio Gianelle}
\affiliation{INFN, Sezione di Padova, I-35131 Padova, Italy.}

\author{Davide Zuliani}
\affiliation{Dipartimento di Fisica e Astronomia ``G. Galilei'', Universit\`a di Padova, I-35131 Padova, Italy}   
\affiliation{INFN, Sezione di Padova, I-35131 Padova, Italy.}

\author{Donatella Lucchesi}
\affiliation{Dipartimento di Fisica e Astronomia ``G. Galilei'', Universit\`a di Padova, I-35131 Padova, Italy}   
\affiliation{INFN, Sezione di Padova, I-35131 Padova, Italy.}

\author{Simone Montangero}
\affiliation{Dipartimento di Fisica e Astronomia ``G. Galilei'', Universit\`a di Padova, I-35131 Padova, Italy}   
\affiliation{INFN, Sezione di Padova, I-35131 Padova, Italy.}
\affiliation{
Padua Quantum Technologies Research Center, Universit\`a degli Studi di Padova, I-35131 Padova, Italy}

\date{\today}

\begin{abstract}

Tensor Networks, a numerical tool originally designed for simulating quantum many-body systems, have recently been applied to solve Machine Learning problems. Exploiting a tree tensor network, we apply a quantum-inspired machine learning technique to a very important and challenging big data problem in high energy physics: the analysis and classification of data produced by the Large Hadron Collider at CERN. In particular, we present how to effectively classify so-called b-jets, jets originating from b-quarks from proton-proton collisions in the LHCb experiment, and how to interpret the classification results. We exploit the Tensor Network approach to select important features and adapt the network geometry based on information acquired in the learning process. Finally, we show how to adapt the tree tensor network to achieve optimal precision or fast response in time without the need of repeating the learning process. These results pave the way to the implementation of high-frequency real-time applications, a key ingredient needed among others for current and future LHCb event classification able to trigger events at the tens of MHz scale. 
\end{abstract}

\maketitle


\begin{center} 
\begin{figure*}[t!]
\includegraphics[width=0.95\textwidth]{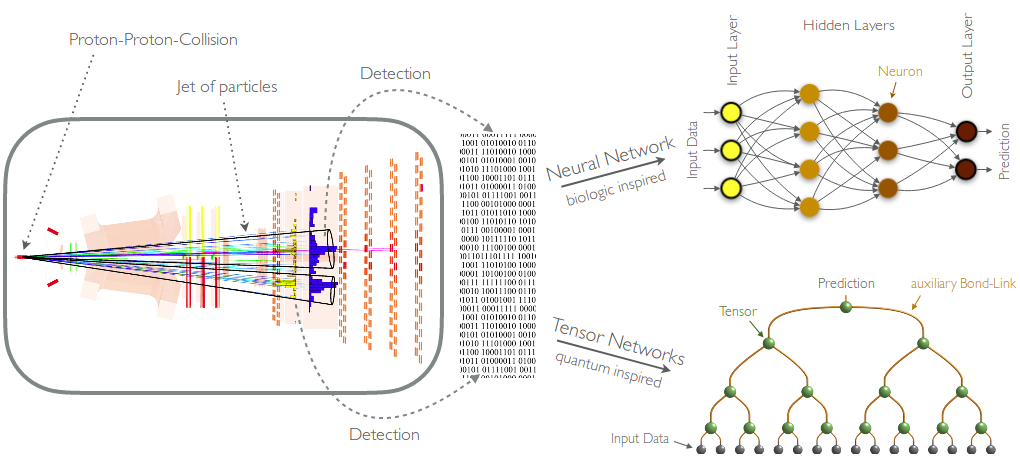}
\captionsetup{justification=centerlast}
\caption{ \label{fig:MLTTN}
Data flow of the Machine Learning analysis for the $b$-jet classification of the LHCb experiment at CERN. After Proton-Proton collisions, $b$- and $\bar{b}$-quarks are created, which subsequently fragment into particle jets (left). The different particles within the jets are tracked by the LHCb detector. Selected features of the detected particle data are used as input for the Machine Learning analysis by NNs and TNs in order to determine the charge of the initial quark (right).
}
\end{figure*}
\end{center}
\section*{Introduction}

Artificial Neural Networks (NN) are a well-established tool for applications in Machine Learning and they are of increasing interest in both research and industry~\cite{bishop1996neural, haykin2009neural, nielsen2015neural, goodfellow2016aaron, LeCun_15, silver2016mastering}. Inspired by biological neural networks, they are able to recognize patterns while processing a huge amount of data. In a nutshell, a NNs describes a functional mapping containing many variational parameters, which are optimised during the training procedure. Recently, deep connections between Machine Learning and quantum physics have been identified and continue to be uncovered~\cite{Carleo_2019}. On one hand, NNs have been applied to describe the behavior of complex quantum many-body systems~\cite{Deng_2017, Nomura_2017, Carleo_2017} while, on the other hand, quantum-inspired technologies and algorithms are taken into account to solve Machine Learning tasks~\cite{Schuld_2018, Das_Sarma_2019, Stoudenmire_2018}.

One particular numerical method originated from quantum physics which has been increasingly compared to NNs are Tensor Networks (TNs)~\cite{TTNvsNN, Chen_2018, levine2017deep}. TNs have been developed to investigate quantum many-body systems on classical computers by efficiently representing the exponentially large quantum wavefunction $|\psi \rangle$ in a compact form and they have proven to be an essential tool for a broad range of applications~\cite{m07, Schollw_ck_2011, sv13, dm16, gqh17, bauls2019simulating, TTNA19, felser2019twodimensional, SimoneBook, Carmen_Ba_uls_2020}. The accuracy of the TN approximation can be controlled with the so-called \textit{bond-dimension} $\chi$, an auxiliary dimension for the indices of the connected local tensors. Recently, it has been shown that TN methods can also be applied to solve Machine Learning (ML) tasks very effectively~\cite{NIPS2016_6211, Stoudenmire_2018, alex2016exponential, levine2017deep, khrulkov2017expressive, Liu_2019,roberts2019tensornetwork}.
Indeed, even though NNs have been highly developed in recent decades by industry and research, the first approaches of Machine Learning with TN yield already comparable results when applied to standard datasets~\cite{NIPS2016_6211, Stoudenmire_2018, glasser2018probabilistic}.
Due to their original development focusing on quantum systems, TNs allow to easily compute quantities such as quantum correlations or entanglement entropy and thereby they grant access to insights on the learned data from a distinct point of view for the application in ML~\cite{levine2017deep, Liu_2019}. Hereafter, we demonstrate the effectiveness of the approach and, more importantly, that it allows introducing algorithms to simplify and explain the learning process, unveiling a pathway to an explainable Artificial Intelligence. As a potential application of this approach, we present a  TN supervised learning of identifying the charge of $b$-quarks (i.e. $b$ or $\bar{b}$) produced in high-energy proton-proton collisions at the Large Hadron Collider (LHC) accelerator at CERN.

In what follows, we first describe the quantum-inspired Tree Tensor Network (TTN) and introduce different quantities that can be extracted from the TTN classifier which are not easily accessible for the biological-inspired Deep NN (DNN), such as correlation functions and entanglement entropy which can be used to explain the learning process and subsequent classifications, paving the way to an efficient and transparent ML tool.
In this regard, we introduce the Quantum-Information Post-learning feature Selection (QuIPS), a protocol that reduces the complexity of the ML model based on the information the single features provide for the classification problem. We then briefly describe the LHCb experiment and its simulation framework, the main observables related to $b$-jets physics, and the relevant quantities for this analysis together with the underlying LHCb data \cite{lhcb_open, lhcb_doi}. 
We further compare the performance obtained by the DNN and the TTN, before presenting the analytical insights into the TTN which, among others, can be exploited to improve future data analysis of high energy problems for a deeper physical understanding of the LHCb data.
Moreover, we introduce the Quantum-Information Adaptive Network Optimisation (QIANO), which adapts the TN representation by reducing the number of free parameters based on the captured information within the TN while aiming to maintain the highest accuracy possible. Therewith, we can optimise the trained TN classifier for a targeted prediction speed without the necessity to relearn a new model from scratch.
\\


Tensor Networks (TNs) are not only a well-established way to represent a quantum wavefunction $|\psi \rangle$, but more general an efficient representation of information as such. In the mathematical context, a TN approximates a high-order tensor by a set of low-order tensors that are contracted in a particular underlying geometry and have common roots with other decompositions, such as the Singular Value Decomposition (SVD) or Tucker decomposition~\cite{Tucker1966}. Among others, some of the most successful TN representations are the Matrix Product State (MPS) - or Tensor Trains~\cite{_stlund_1995, Schollw_ck_2011, ttrains, NIPS2016_6211}, the Tree Tensor Network (TTN) - or Hierarchical Tucker decomposition~\cite{TTN14,HTdec,Liu_2019}, and the Projected Entangled Pair States (PEPS)~\cite{verstraete2004renormalization, Or_s_2014}.

For a supervised learning problem, a TN can be used as the \textit{weight tensor} $W$~\citep{NIPS2016_6211,Stoudenmire_2018,Liu_2019}, a high-order tensor which acts as classifier for the input data $\{{\bf x}\}$: Each sample $\bf x$ is encoded by a \textit{feature map} $\Phi({\bf x})$ and subsequently classified by the \textit{weight tensor} $W$. The final confidence of the classifier for a certain class labeled by $l$ is given by the probability
\be
\mathcal{P}_{l}({\bf x}) = W_l\cdot \Phi({\bf x}) ~.
\ee

In the following, we use a TTN $\Psi$ to represent $W$ (see Fig.~\ref{fig:MLTTN}, bottom right) which can be described as a contraction of its $N$ hierarchically connected local tensors $T_{\{\chi\}}$
\be
\Psi = \sum_{\chi } T_{l,\chi_1, \chi_2}^{[1]} \prod_{\eta=2}^{N} T_{\chi_n,\chi_{2n}, \chi_{2n+1}}^{[\eta]}
\ee
where $n\in[1,N]$.
Therefore, 
we can interpret the TTN classifier $\Psi$ as well as a set of quantum many-body wavefunctions $|\psi_l\rangle$ - one for each of the class labels $l$ (see Supplementary Methods). For the classification, we represent each sample $x$ by a product state $\Phi(x)$. Therefore, we map each feature $x_i\in {\bf x}$ into a quantum spin by choosing the \textit{feature map} $\Phi({\bf x})$ as a Kronecker product of $N+1$ \textit{local feature maps} 
\be
\Phi^{[i]}(x_i) = \left[ \cos{\left( \frac{\pi x_i'}{2} \right)},\sin{\left( \frac{\pi x_i'}{2} \right)} \right]
\label{eq:feature_map}
\ee
where $x_i'\equiv x_i/x_{i,\text{max}}\in [0,1]$ is the re-scaled value with respect to the maximum $x_{i,\text{max}}$ within all samples of the training set. 

Accordingly, we classify a sample $x$ by computing the overlap $\langle \Phi(x) | \psi_l \rangle$ for all labels $l$ with the product state $\Phi(x)$ resulting in the weighted probabilities
\be
\mathcal{P}_l = \frac{|\langle \Phi(x)| \psi_l\rangle|^2}{\sum_{l}|\langle \Phi(x)| \psi_l\rangle|^2}
\ee
for each class. We point out, that we can encode the input data in different non-linear feature maps as well (see Supplementary Notes).\\

One of the major benefits of Tensor Networks in quantum mechanics is the accessibility of information within the network. They allow to efficiently measure information quantities such as entanglement entropy and correlations. Based on these quantum-inspired measurements, we here introduce the QuIPS protocol for the TN application in Machine Learning, which exploits the information encoded and accessible in the TN in order to rank the input features according to their importance for the classification.

In information theory, entropy as such is a measure of the information content inherent in the possible outcomes of variables, such as e.g. a classification~\citep{6773024,6773067,NiC00}. In TNs such information content can be assessed by means of the entanglement entropy $S$ which describes the shared information between TN bipartitions. The entanglement $S$ is measured via the Schmidt decomposition, that is, decomposing  $|\psi\rangle$ into two bipartitions $|\psi^A_\alpha \rangle$ and $|\psi^B_\alpha \rangle$~\cite{NiC00} such that
\be
\Psi = \sum_\alpha^\chi \lambda_\alpha |\Psi^A_\alpha\rangle \otimes |\Psi^B_\alpha\rangle ,
\ee
where $\lambda_\alpha$ are the Schmidt-coefficients (non-zero, normalised singular values of the decomposition). The entanglement entropy is then defined as $S=-\sum_\alpha \lambda_\alpha^2 \ln{\lambda_\alpha^2}$. Consequently, the minimal entropy $S=0$ is obtained only if we have one single non-zero singular value $\lambda_1=1$. In this case, we can completely separate the two bipartitions as they share no information. On the contrary, higher $S$ means that information is shared among the bipartitions. 

In the Machine Learning context, the entropy can be interpreted as follows: If the features in one bipartition provide no valuable information for the classification task, the entropy is zero. On the contrary, $S$ increases the more information between the two bipartitions are exploited. This analysis can be used to optimize the learning procedure: whenever $S=0$, the feature can be discarded with no loss of information for the classification. Thereby, a second model with fewer features and fewer tensors can be introduced. This second, more efficient model results in the same predictions in less time. On the contrary,  a high bipartition entropy highlights which feature - or combination of features - are important for the correct predictions. 

The second set of measurements we take into account are the correlation functions 
\be
C^l_{i,j} = \frac{\langle \psi_{l}| \sigma_i^z \sigma_j^z | \psi_l \rangle}{\langle \psi_{l}| \psi_l \rangle}
\ee
for each pair of features (located at site $i$ and $j$) and for each class $l$. The correlations offer an insight into the possible relation among the information that the two features provide. In case of maximum correlation or anti-correlation among them for all classes $l$, the information of one of the features can be obtained by the other one (and vice versa), thus one can be neglected. In case of no correlation among them, the two features may provide fundamentally different information for the classification. 
The correlation analysis allows pinpointing if two features give independent information. However, the correlation itself - in contrast to the entropy - does not tell if this information is important for the classification.

In conclusion, based on the previous insights, namely {\it (i)}
    a low entropy of a feature bipartition signals that one of the two bipartitions can be discarded, providing negligible loss of information and
    {\it (ii)} if two features are completely (anti-)correlated we can neglect at least one of them, the QuIPS enables to filter out the most valuable features for the classification.
\\


Nowadays, in particle physics, ML is widely used for the classification of jets, \textit{i.e.} streams of particles produced by the fragmentation of quarks and gluons.
The jet substructure can be exploited to solve such classification problems \cite{jet_substructure}.
ML techniques have been proposed to identify boosted, hadronically decaying top quarks \cite{top_tagger}, or to identify the jet charge \cite{jet_ml}.
The ATLAS and CMS collaborations developed ML algorithms in order to identify jets generated by the fragmentation of $b$-quarks \cite{atlas_ml1, atlas_ml2, cms_ml}: a comprehensive review on ML techniques at the LHC can be found in \cite{jet_lhc}.

The LHCb experiment in particular is, among others, dedicated to the study of the physics of $b$- and $c$-quarks produced in proton-proton collisions. Here, ML methods have been introduced recently
for the discrimination between $b$- and $c$-jets by using Boosted Decision Tree classifiers~\cite{lhcb_tagging}. However, a crucial topic for the LHCb experiment, which is yet unexploited by ML, is the identification of the charge of a $b$-quark, \textit{i.e.} discriminating between a $b$ or $\bar{b}$. Such identification can be used in many physics measurements, and it is the core of the determination of the charge asymmetry in ${b}$-pairs production, a quantity sensitive to physics beyond the Standard Model~\cite{asy_np}.
Whenever produced in a scattering event, $b$-quarks have a short lifetime as free particles; indeed, they manifest themselves as bound states (hadrons) or as narrow cones of particles produced by the hadronization (jets). In the case of the LHCb experiment, the $b$-jets are detected by the apparatus located in the forward region of proton-proton collisions (see Fig.~\ref{fig:MLTTN}, left)~\cite{lhcb_det}. The LHCb detector includes a particle identification system that distinguishes different types of charged particles within the jet, and a high-precision tracking system able to measure the momentum of each particles~\cite{lhcb_perf}. Still, the separation between $b$- and $\bar{b}$-jets is a highly difficult task because the $b$-quark fragmentation produce dozens of particles via non-perturbative Quantum Chromodynamics processes, resulting in non-trivial correlations between the jet particles and the original quark.

The algorithms used to identify the charge of the $b$-quarks based on information on the jets
are called tagging methods. The tagging algorithm performance is typically quantified with the \textit{tagging power} $\epsilon_{tag}$, representing the effective fraction of jets that contribute to the statistical uncertainty in an asymmetry measurement~\cite{tag_pow1, tag_pow2}. In particular, the tagging power $\epsilon_{tag}$ takes into account the efficiency $\epsilon_{eff}$ (the fraction of jets for which the classifier takes a decision) and the prediction accuracy $a$ (the fraction of correctly classified jets among them) as follows: 
\be
\epsilon_{tag} = \epsilon_{eff} \cdot (2a-1)^2 ~.
\ee
To date, the muon tagging method gives the best performance on the $b$- vs $\bar{b}$-jet discrimination using the dataset collected in the LHC Run I~\cite{lhcb_asy}:
here, the muon with the highest momentum in the jet is selected, and its electric charge is used to decide on the $b$-quark charge.

For the ML application, we now formulate the identification of the $b$-quark charge in terms of a supervised learning problem. As described above, we implemented a TTN as a classifier and applied it to the LHCb problem analysing its performance. Alongside, a DNN analysis is performed to the best of our capabilities, and both algorithms are compared with the muon tagging approach.
Both the TTN and the DNN, use as input for the supervised learning $16$ features of the jet substructure from the official simulation data released by the LHCb collaboration \cite{lhcb_open, lhcb_doi}.
The $16$ features are determined as follows: the muon with the highest $p_\mathrm{T}$ among all other detected muons in the jet is selected and the same is done for the highest $p_\mathrm{T}$ kaon, pion, electron, and proton, resulting in  5 different selected particles. For each particle, three observables are considered: (i) The momentum relative to the jet axis ($p^{rel}_{\mathrm{T}}$), (ii) the particle charge ($q$), and (iii) the distance between the particle and the jet axis ($\Delta R$), for a total of  $5 \times 3$ observables. 
If a particle type is not found in a jet, the related features are set to $0$.
The $16$th feature is the total jet charge $Q$, defined as the weighted average of the particles charges $q_i$ inside the jet, using the particles $p^{rel}_{\mathrm{T}}$ as weights:
\be
Q =  \frac{\sum_i (p^{rel}_{\mathrm{T}})_i q_i}{\sum_i (p^{rel}_{\mathrm{T}})_i} ~.
\ee

\section*{Results}
\subsection*{Analysis framework}
In the following, we present the jet classification performance for the TTN and the DNN applied to the LHCb dataset, also comparing both ML techniques with the muon tagging approach.
For the DNN we use an optimized network with three hidden layers of 96 nodes (see Supplementary Methods for details). Hereafter, we aim to compare the best possible performance of both approaches therefore, we optimised the hyper-parameters of both methods in order to obtain the best possible results from each of them, TTN and DNN. Therefore, we split the dataset of about $700k$ events (samples) into two sub-sets: $60\%$ of the samples are used in the training process while the remaining $40\%$ are used as test set to evaluate and compare the different methods. For each event prediction after the training procedure, both ML models output the probability $\mathcal{P}_{b}$ to classify the event as a jet generated by a $b$- or a $\bar{b}$-quark.  A threshold $\Delta$ around $\mathcal{P}_{b}=0.5$ is then defined in which we classify the quark as unknown in order to optimise the overall tagging power $\epsilon_{tag}$.

\subsection*{Jet classification performance }\label{sec:results}
We obtain similar performances in terms of the raw prediction accuracy applying both ML approaches after the training procedure on the test data: the TTN takes a decision on the charge of the quark in $\epsilon_{eff}^{\text{TTN}}=54.5\%$ of the cases with an overall accuracy of $a^{\text{TTN}}=70.56\%$, while the DNN decides in $\epsilon_{eff}^{\text{DNN}}=55.3\%$ of the samples with $a^{\text{DNN}}=70.49\%$. We checked both approaches for biases in physical quantities to ensure that both methods are able to properly capture the physical process behind the problem and thus that they can be used as valid tagging methods for LHCb events (see Supplementary Methods).

\begin{figure}[ht]
\centering
\begin{minipage}{.225\linewidth}
  \centering
  \includegraphics[width=\linewidth]{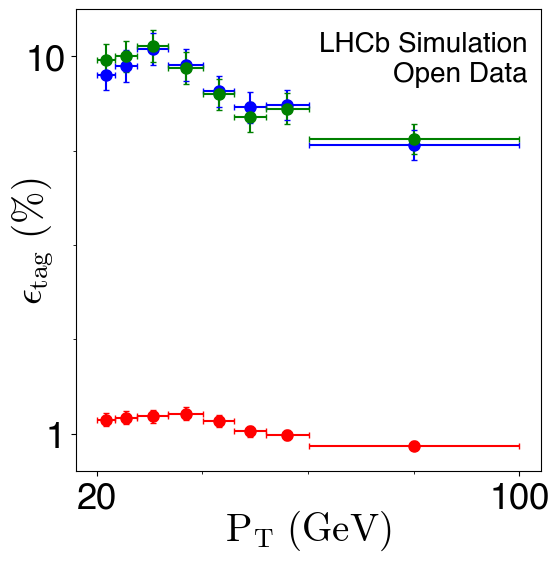}
  \subcaption{}
  \label{fig:Tagging}
\end{minipage}%
\begin{minipage}{.225\linewidth}
  \centering
  \includegraphics[width=\linewidth]{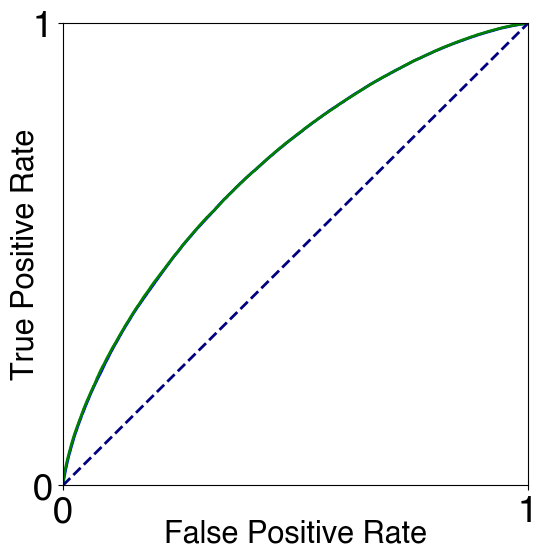}
  \subcaption{}
  \label{fig:roc}
\end{minipage}
\begin{minipage}{.25\linewidth}
  \centering
  \includegraphics[width=\linewidth]{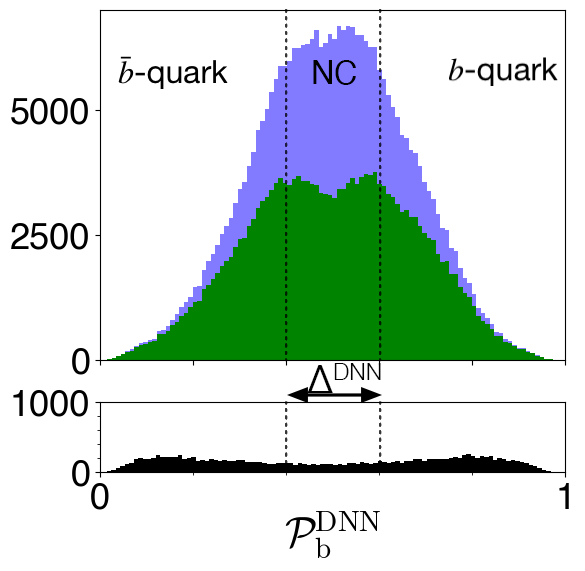}
  \subcaption{}
  \label{fig:probDNN}
\end{minipage}%
\begin{minipage}{.25\linewidth}
  \centering
  \includegraphics[width=\linewidth]{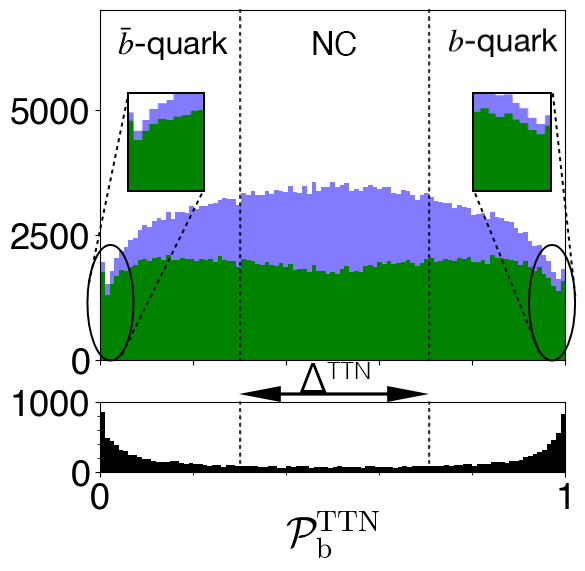}
  \subcaption{}
  \label{fig:probTTN}
\end{minipage}
\caption{Comparison of the DNN and TNN analysis: (a) Tagging power for the DNN (green), TTN (blue) and the muon tagging (red), (b) ROC curves for the DNN (green) and the TTN (blue, but completely covered by DNN), compared with the \emph{line of no-discrimination} (dotted navy-blue line), (c) probability distribution for the DNN, and (d) for the TTN. In the two distributions (c)+(d), the correctly classified events (green) are shown in the total distribution (light blue). Below, in black all samples where a muon was detected in the jet.}
\end{figure}

In Fig.~\ref{fig:Tagging} we present the tagging power of the different approaches as a function of the jet transverse momentum $p_\mathrm{T}$. Evidently, both Machine Learning methods perform significantly better than the muon tagging approach for the complete range of jet transverse momentum $p_T$, while the TTN and DNN display comparable performances within the statistical uncertainties.

In Figs.~\ref{fig:probDNN} and ~\ref{fig:probTTN} we present the histograms of the confidences for predicting a $b$-flavored jet for all samples in the test data set for the DNN and the TTN respectively. Interestingly, even though both approaches give similar performances in terms of overall precision and tagging power, the prediction confidences are fundamentally different. For the DNN, we see a Gaussian-like distribution with, in general, not very high confidence for each prediction. Thus, we obtain less correct predictions with high confidences, but at the same time, fewer wrong predictions with high confidences compared to the TTN predictions. On the other hand, the TTN displays a flatter distribution including more predictions - correct and incorrect - with higher confidence. Remarkably though, we can see peaks for extremely confident predictions (around 0 and around 1) for the TTN. These peaks can be traced back to the presence of the muon; noting that the charge of which is a well-defined predictor for a jet generated by a $b$-quark. The DNN lacks these confident predictions exploiting the muon charge.
Further, we mention that using different cost functions for the DNN, i.e. cross-entropy loss function and the Mean Squared Error, lead to similar results (see Supplementary Methods).

Finally, in Fig.~\ref{fig:roc} we present the Receiving Operator Characteristic (ROC) curves for the TTN and the DNN together with the \emph{line of no-discrimination}, which represents a randomly guessing classifier: the two ROC curves for TTN and DNN are perfectly coincident, and the Area Under the Curve (AUC) for the two classifiers is the almost same ($AUC^{TTN}=0.689$ and $AUC^{DNN}=0.690$).
The graph illustrates the similarity in the outputs between TTN and DNN despite the different confidence distributions. This is further confirmed by a Pearson correlation factor of $r=0.97$ between the outputs of the two classifiers.

In conclusion, the two different approaches result in similar outcomes in terms of prediction performances. However, the underlying information used by the two discriminators is inherently different. For instance, the DNN predicts more conservatively, in the sense that the confidences for each prediction tend to be lower compared with the TTN. Additionally, the DNN does not exploit the presence of the muon as strongly as the TTN, even though the muon is a good predictor for the classification.

\subsection*{Exploiting insights into the data with TTN}\label{sec:insights}
As previously mentioned, the TTN analysis allows to efficiently measure the captured correlations and the entanglement within the classifier. These measurements give insight into the learned data and can be exploited via QuIPS to identify the most important features typically used for the classifications.

\begin{center} 
\begin{figure}[ht]
\centering
\begin{minipage}{.25\linewidth}
  \centering
  \includegraphics[width=\linewidth]{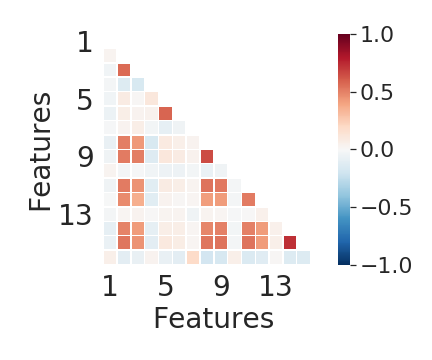}
  \subcaption{}
  \label{fig:Corr}
\end{minipage}%
\begin{minipage}{.25\linewidth}
  \centering
  \includegraphics[width=\linewidth]{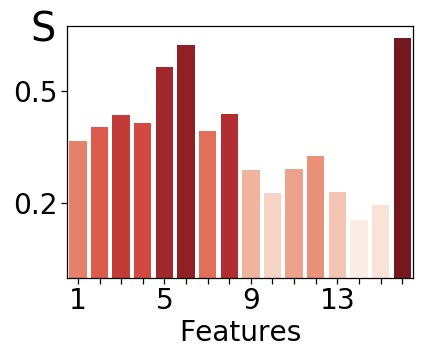}
  \subcaption{}
  \label{fig:Entr}
\end{minipage}
\begin{minipage}{.225\linewidth}
  \centering
  \includegraphics[width=\linewidth]{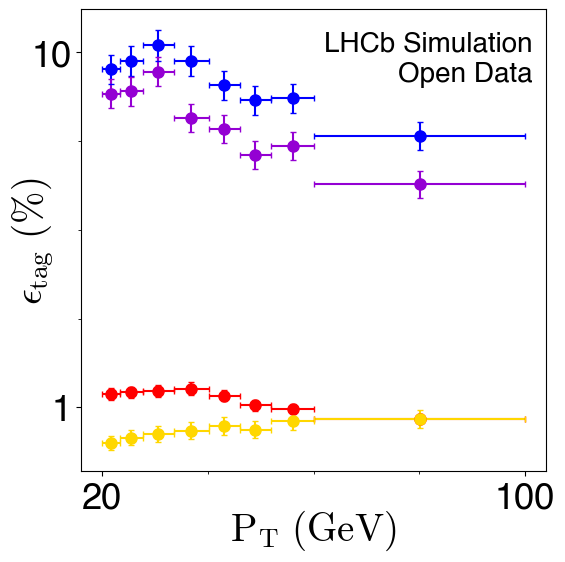}
  \subcaption{}
  \label{fig:QR_TagPower}
\end{minipage}%
\begin{minipage}{.225\linewidth}
  \centering
  \includegraphics[width=\linewidth]{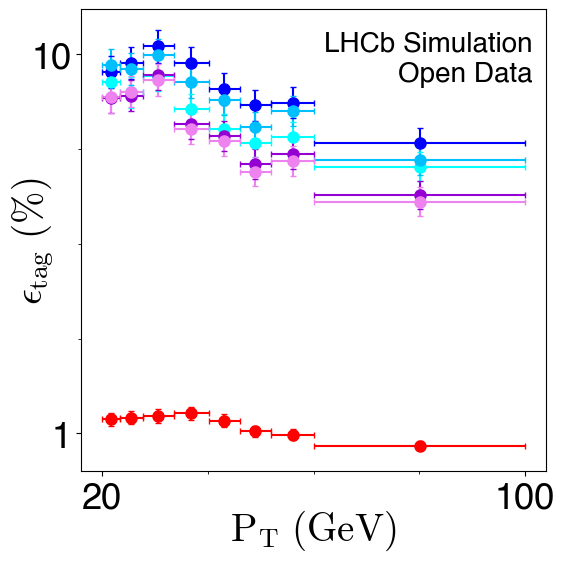}
  \subcaption{}
  \label{fig:Trunc_TagPower}
\end{minipage}
\caption{Exploiting the information provided by the learned TTN classifier: (a) Correlations between the $16$ input features (blue for anti-correlated, white for uncorrelated, red for correlated). The numbers indicate $q$, $p_T^{rel}$, $\Delta R$ of the muon (1-3), kaon (4-6), pion (7-9), electron (10-12), proton (13-15) and the jet charge $Q$ (16). 
(b) Entropy of each feature as the measure for the information provided for the classification. (c) Tagging power for learning on all features (blue), the best 8 proposed by QuIPS exploiting insights from (a)+(b) (magenta), the worst 8 (yellow) and the muon tagging (red). (d) Tagging power for decreasing bond dimension truncated after training: The complete model (blue shades for $\chi=100$, $\chi=50$, $\chi=5$), for using the QuIPS best 8 features only (violet shades for $\chi=16$, $\chi=5$), and the muon tagging (red).}
\end{figure}
\end{center}

In Fig.~\ref{fig:Corr} we present the correlation analysis allowing us to pinpoint if two features give independent information. For both labels ($l=b,\bar{b}$) the results are very similar, thus in Fig.~\ref{fig:Corr} we present only $l=b$. We see among others that the momenta $p_T^{rel}$ and distance $\Delta R$ of all particles are correlated except for the kaon. Thus this particle provides information to the classification which seems to be independent of the information gained by the other particles. However, the correlation itself does not tell if this information is important for the classification. Thus, we compute the entanglement entropy $S$ of each feature, as reported in Fig.~\ref{fig:Entr}. Here, we conclude that the features with the highest information content are the total charge and $p_T^{rel}$ and distance $\Delta R$ of the kaon. Driven by these insights, we employ the QuIPS to discard half of the features by selecting the 8 most important ones: 
i.-iii. charge, momenta, and distance of the muon; iv.-vi. charge, momenta, and distance of the kaon; vii. charge of the pion; viii. total detected charge. To test the QuIPS performance, we compared it with an independent but more time-expensive analysis on the importance of the different particle types: the two approaches perfectly matched. 
Further, we studied two new models, one composed of the 8 most important features proposed by the QuIPS, and, for comparison, another with the 8 discarded features. In Fig.~\ref{fig:QR_TagPower} we show the tagging power for the different analyses with the complete 16-sites (model $M_{16}$), the best 8 ($B_8$), the worst 8 ($W_8$) and the muon tagging. Remarkably, we see that
the models $M_{16}$ and $B_8$ give 
comparable results, while model $W_8$
results are even worse than the classical approach. These performances are confirmed by the prediction accuracy of the different models: While only less than $1\%$ of accuracy is lost from $M_{16}$ to $B_8$, the accuracy of the model $W_8$ drastically drops to around $52\%$ - that is, almost random predictions. Finally, in this particular run, the model $B_8$ has been trained $4.7$ times faster with respect to model $M_{16}$ and predicts $5.5$ times faster as well (The actual speed-up depends on the bond-dimension and other hyperparameters).

\begin{table*}[ht]
    \centering
    \begin{tabular}{c||c c c |c c c}
        & \multicolumn{3}{c|}{Model $M_{16}$ (incl. all 16 features)} & \multicolumn{3}{|c}{Model $B_8$ (best 8 features determined by QuIPS)} \\
        $\bf \chi$ & Prediction time & Accuracy & Free parameters & Prediction time & Accuracy  & Free parameters \\
        \hline
        $\bf 200$  & $345\,\mu$s & $70.27~\%~(63.45~\%)$ & $51501$ & -&- & -\\
        $\bf 100$  & $178\,\mu$s& $70.34~\%~(63.47~\%)$ & $25968$ &-&-&-\\
        $\bf 50$ & $105\,\mu$s & $70.26~\%~(63.47~\%)$ & $13214$ &-&-&-\\
        $\bf 20$ & $62\,\mu$s & $70.31~\%~(63.46~\%)$ & $5576$ &- &- &-\\
        $\bf 16$ & - & - & - & $19\,\mu$s &$69.10~\%~(62.78~\%)$ &$264$\\
        $\bf 10$ & $40\,\mu$s & $70.36~\%~(63.44~\%)$ & $1311$ & $19\,\mu$s &$69.01~\%~(62.78~\%)$ & $171$\\
        $\bf 5$ & $37\,\mu$s & $69.84~\%~(62.01~\%)$ & $303$ & $19\,\mu$s &$69.05~\%~(62.76~\%)$ & $95$
    \end{tabular}
    \caption{ Prediction time, accuracy with (and without) applied cuts $\Delta$ and number of free parameters of the TTN for different bond dimension $\chi$ when we reduce the TTN model with QIANO, both for the complete 16 (left) and the QuIPS reduced 8 features (right). For the model $M_{16}$ with all 16 features (left), we trained the TTN with $\chi=200$ and truncate from there while for the reduced model $B_8$ (right), the original bond-dimension was $\chi=16$ (being the maximum $\chi$ in this subspace).}
    \label{tab:table_QIANO}
\end{table*}

A critical point of interest in real-time ML applications is the prediction time.  For example, in the LHCb Run 2 data-taking, the high-level software trigger takes a decision approximately every $1 \ \mu$s~\cite{lhcb_perf} and shorter latencies are expected in future Runs. Consequently, with the aid of the QuIPS protocol, we can efficiently reduce the prediction computational time while maintaining a comparable high prediction power.
However, with TTNs, we can undertake an even further step to reduce the prediction time by reducing the bond dimension $\chi$ after the training procedure. Here, we introduce the \textit{Quantum information Adaptive Network Optimization} (QIANO) performing this truncation by means of the well-established SVD for TN~\cite{SimoneBook,Schollw_ck_2011,TTNA19} in a way ensuring to introduce the least infidelity possible. In other words, QIANO can adjust the bond dimension $\chi$ to achieve a targeted prediction time while 
keeping the 
prediction accuracy reasonably high. We stress that this can be done without relearning a new model, as would be the case with NN.

Finally, we apply QuIPS and QIANO to reduce the information in the TTN in an optimal way for a targeted balance between prediction time and accuracy. In Fig.~\ref{fig:Trunc_TagPower} we show the tagging power taking the original TTN and truncate it to different bond-dimensions $\chi$. We can see, that even though we compress quite heavily, the overall tagging power does not change significantly. In fact, we only drop about $0.03\%$ in the overall prediction accuracy, while at the same time improving the average prediction time from $345\, \mu$s to $37\, \mu$s (see Tab.~\ref{tab:table_QIANO}). Applying the same idea to the model $B_8$ we can reduce the average prediction time effectively down to $19\, \mu$s on our machines, a performance compatible with current real-time classification rate. 

\section*{Discussion}\label{sec:conclusion}

We analysed an LHCb dataset for the classification of $b$- and $\bar{b}$-jets with two different ML approaches, a DNN and a TTN. We showed that we obtained with both techniques a tagging power about one order of magnitude higher than the classical muon tagging approach, which up to date is the best-published result for this classification problem. We pointed out that, even though both approaches result in similar tagging power, they treat the data very differently. In particular, TTN effectively recognises the importance of the presence of the muon as a strong predictor for the jet classification. Here, we point out that we only used a conjugate gradient descent for the optimisation of our TTN classifier. Deploying more sophisticated optimisation procedures which have already been proven to work for Tensor Trains, such as stochastic gradient descent~\cite{torchmps} or Riemannian optimisation~\cite{alex2016exponential}, may further improve the performance (in both time and accuracy) in future applications.

We further explained the crucial benefits of the TTN approach over the DNNs, namely (i) the ability to efficiently measuring correlations and the entanglement entropy, and (ii) the power of compressing the network while keeping a high amount of information (to some extend even lossless compression). We showed how the former quantum-inspired measurements help to set up a more efficient ML model: in particular, by introducing an information-based heuristic technique, we can establish the importance of single features based on the information captured within the trained TTN classifier only. Using this insight, we introduced the QuIPS, which can significantly reduce the model complexity by discarding the least-important features maintaining high prediction accuracy.
This selection of features based on their informational importance for the trained classifier is one major advantage of TNs targeting to effectively decrease training and prediction time. Regarding the latter benefit of the TTN, we introduced the QIANO, which allows to decrease the TTN prediction time by optimally decreasing its representative power based on information from the quantum entropy,  introducing the least possible infidelity. In contrast to DNNs, with the QIANO we do not need to set up a new model and train it from scratch, but we can optimise the network post-learning adaptively to the specific conditions, e.g., the used CPU or the required prediction time of the final application.

Finally, we showed that using QuIPS and QIANO we can effectively compress the trained TTN to target a given prediction time. 
In particular, we decreased our prediction times from $345\mu s$ to $19 \mu s$. We stress that, while we only used one CPU for the predictions, in future application we might obtain a speed-up from $10$ to $100$ times by parallelising the tensor contractions on GPUs~\cite{milsted2019tensornetwork}. Thus, 
we are confident that it is possible to reach a MHz prediction rate while still obtaining results significantly better than the classical muon tagging approach.
Here, we also point out that, for using this algorithm on the LHCb real-time data acquisition system, it would be necessary to develop custom electronic cards like FPGAs, or GPUs with an optimized architecture. Such solutions should be explored in the future.

Given the competitive performance of the presented TTN method at its application in high energy physics, we envisage a multitude of possible future applications in high-energy experiments at CERN and in other fields of science. Future applications of our approach in the LHCb experiment may include the discrimination between $b$-jets, $c$-jets and light flavour jets~\cite{lhcb_tagging}. A fast and efficient real-time identification of $b$- and $c$-jets can be the key point for several studies in high energy physics, ranging from the search for the rare Higgs boson decay in two $c$-quarks, up to the search for new particles decaying in a pair of heavy-flavour quarks ($b\bar{b}$ or $c \bar{c}$). 

\section*{Methods}

\begin{center} 
\begin{figure}[ht]
\includegraphics[width=0.35\textwidth]{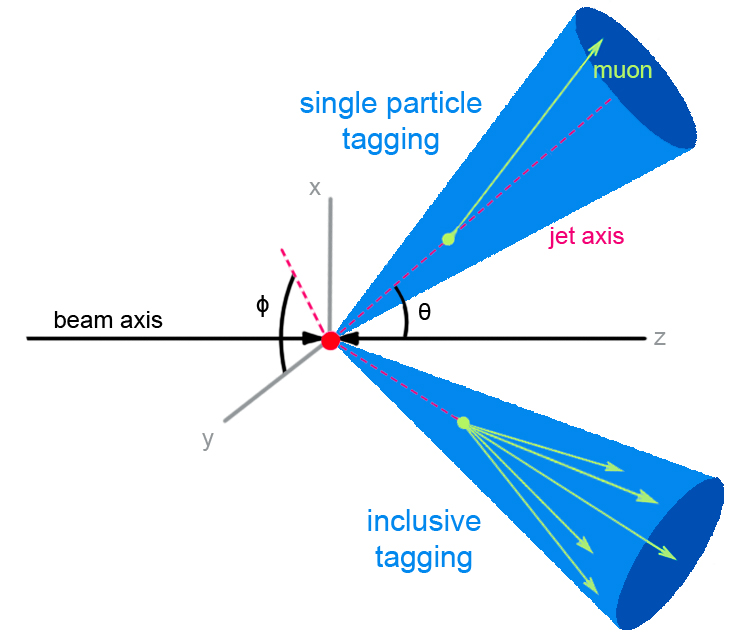}
\caption{ \label{fig:Btagging}
Illustrative sketch showing an LHCb experiment and the two possible tagging algorithms: a single particle tagging algorithm, exploiting information coming from one single particle (muon), and the inclusive tagging algorithm which exploits the information on all the jet constituents.
}
\end{figure}
\end{center}

\paragraph*{\textbf{LHCb particle detection ---}} LHCb is fully instrumented in the phase space region of proton-proton collisions defined by the pseudo-rapidity ($\eta$) range [2,5], with $\eta$ defined as 
\be
\eta = -\mathrm{log}\left[\mathrm{tan}\left(\frac{\theta}{2}\right)\right] ~,
\ee
where $\theta$ is the angle between the particle momentum and the beam axis (see Fig.~\ref{fig:Btagging}). The direction of particles momenta can be fully identified by $\eta$ and by the azimuthal angle $\phi$, defined as the angle in the plane transverse to the beam axis. The projection of the momentum in this plane is called transverse momentum ($p_{\mathrm{T}}$). The energy of charged and neutral particles is measured by electromagnetic and hadronic calorimeters. In the following, we work with physics natural units.

At LHCb jets are reconstructed using a Particle Flow algorithm~\cite{aleph_perf} for charged and neutral particles selection and using the anti-$k_t$ algorithm~\cite{anti_kt} for clusterization. The jet momentum is defined as the sum of the momenta of the particles that form the jet, while the jet axis is defined as the direction of the jet momentum. Most of the particles that form the jet are contained in a cone of radius $\Delta R=\sqrt{(\Delta \eta)^2+(\Delta \phi)^2}=0.5$, where $\Delta \eta$ and $\Delta \phi$ are respectively the pseudo-rapidity difference and the azimuthal angle difference between the particles momenta and the jet axis. For each particle inside the jet cone, the momentum relative to the jet axis ($p^{rel}_{\mathrm{T}}$) is defined as the projection of the particle momentum in the plane transverse to the jet axis.
\\

\paragraph*{\textbf{LHCb Dataset ---}}
Differently from other ML performance analyses, the dataset used in this paper has been prepared specifically for this LHCb classification problem, therefore baseline ML models and benchmarks on it do not exist. In particle physics, features are strongly dependent on the detector considered (\emph{i.e.} different experiments may have a different response on the same physical object) and for this reason the training has been performed on a dataset that reproduces the LHCb experimental conditions, in order to obtain the optimal performance with this experiment.

The LHCb simulation datasets used for our analysis are produced with a Monte Carlo technique using the framework GAUSS~\cite{gauss}, which makes use of PYTHIA 8~\cite{pythia} to generate proton-proton interactions and jet fragmentation and uses EvtGen~\cite{evtgen} to simulate $b$-hadrons decay. The GEANT4 software~\cite{geant4_1,geant4_2} is used to simulate the detector response, and the signals are digitized and reconstructed using the LHCb analysis framework.

The used dataset contains $b$ and $\bar{b}$-jets produced in proton-proton collisions at a center-of-mass energy of $13\,$TeV~\cite{lhcb_open, lhcb_doi}. Pairs of $b$-jets and $\bar{b}$-jets are selected by requiring a jet $p_{\mathrm{T}}$ greater than 20 GeV and $\eta$ in the range [2.2,4.2] for both jets.\\

\paragraph*{\textbf{Muon tagging ---}}
LHCb measured the $b \bar{b}$ forward-central asymmetry using the dataset collected in the LHC Run I~\cite{lhcb_asy} using the muon tagging approach:
In this method, the muon with the highest momentum in the jet cone is selected, and its electric charge is used to decide on the $b$-quark charge.
In fact, if this muon is produced in the original semi-leptonic decay of the $b$-hadron, its charge is totally correlated with the $b$-quark charge.
Up to date, the muon tagging method gives the best performance on the $b$- vs $\bar{b}$-jet discrimination.
Although this method can distinguish between $b$- and $\bar{b}$-quark with good accuracy, its efficiency is low as it is only applicable on jets where a muon is found and it is intrinsically limited by the $b$-hadrons branching ratio in semi-leptonic decays. Additionally, the muon tagging may fail in some scenarios, where the selected muon is produced not by the decay of the $b$-hadron but in other decay processes. In these cases, the muon may not be completely correlated with the $b$-quark charge.\\

\paragraph*{\textbf{Machine Learning approaches ---}}
We train the TTN and analyse the data with different bond dimensions $\chi$. The auxiliary dimension $\chi$ controls the number of free parameters within the variational TTN ansatz. While the TTN is able to capture more information from the training data with increasing bond dimension $\chi$, choosing $\chi$ too large may lead to overfitting and thus can worsen the results in the test set. For the DNN we use an optimized network with three hidden layers of 96 nodes (see Supplementary Methods for details).

For each event prediction, both methods give as output the probability $\mathcal{P}_{b}$ to classify a jet as generated by a $b$- or a $\bar{b}$-quark. This probability (\textit{i.e.} the confidence of the classifier) is normalized in the following way: for values of probability $\mathcal{P}_{b}>0.5$ ($\mathcal{P}_{b}<0.5$) a jet is classified as generated by a $b$-quark ($\bar{b}$-quark), with an increasing confidence going to $\mathcal{P}_{b}=1$ ($\mathcal{P}_{b}=0$). Therefore a completely confident classifier returns a probability distribution peaked at $\mathcal{P}_{b}=1$ and $\mathcal{P}_{b}=0$ for jets classified as generated by $b$- and $\bar{b}$-quark respectively.

We introduce a threshold $\Delta$ symmetrically around the prediction confidence of $\mathcal{P}_{b}=0.5$ in which we classify the event as unknown. We optimise the cut on the predictions of the classifiers (\textit{i.e.} their confidences) to maximise the tagging power for each method based on the training samples. In the following analysis we find $\Delta^{\text{TTN}} = 0.40$ ($\Delta^{\text{DNN}} = 0.20$) for the TTN (DNN). Thereby, we predict for the TTN (DNN) a $b$-quark with confidences $\mathcal{P}_{b}>C^{\text{TTN}}=0.70$ ($\mathcal{P}_{b}>C^{\text{DNN}}=0.60$), a $\bar{b}$-quark with confidences $\mathcal{P}_{b}<0.30$ ($\mathcal{P}_{b}<0.40$) and no prediction for the range in between (see Fig.~\ref{fig:probDNN} and Fig.~\ref{fig:probTTN}).\\

\subsection*{Data availability}
This paper is based on data obtained by the LHCb experiment, but is analyzed independently, and has not been reviewed by the LHCb collaboration.
The data are available in the official LHCb open data repository \cite{lhcb_open, lhcb_doi}. 

\subsection*{Code availability}
The software code used for the analysis of the Deep Neural Network can be freely acquired when contacting gianelle@pd.infn.it and it is permitted to use it for any kind of private or commercial usage including modification and distribution without any liabilities or warranties. The software code for the TTN analysis is currently not available for public use. For more information, please contact timo.felser@physik.uni-saarland.de.

\subsection*{Acknowledgments}
We are very grateful to Konstantin Schmitz for valuable comments and discussions on the Machine Learning comparison. We thank Miles Stoudenmire for fruitful discussions on the application of the Tensor Networks Machine Learning code.

This work is partially supported by the Italian PRIN 2017 and Fondazione CARIPARO, the Horizon 2020 research and innovation programme under grant agreement No 817482 (Quantum Flagship - PASQuanS) and the QuantERA projects QTFLAG and QuantHEP. We acknowledge computational resources by CINECA and the Cloud Veneto. 

The work is partially supported by the German Federal Ministry for Economic Affairs and Energy (BMWi) and the European Social Fund (ESF) as part of the EXIST program under the project \textit{Tensor Solutions}.

We acknowledge the LHCb Collaboration for the valuable help and the Istituto Nazionale di Fisica Nucleare and the Department of Physics and Astronomy of the University of Padova for the support.

\subsection*{Author contributions}
Conceptualization (TF, DL, SM); Data Analysis (DZ, LS, TF, MT, AG); Funding Acquisition (DL, SM); Investigation (DZ, LS, TF, SM); Methodology (TF, SM); Tensor Network Software Development (TF using private resources); Validation (DZ, LS, TF, MT); Writing – original draft (TF, SM); Writing - review \& editing (all authors).

\subsection*{Competing interests}
The authors declare no competing interests.


\section*{Figure Legends}

\paragraph*{Figure 1 --\textbf{High energy machine learning, NN vs. TTN}}
Data flow of the Machine Learning analysis for the $b$-jet classification of the LHCb experiment at CERN. After Proton-Proton collisions, $b$- and $\bar{b}$-quarks are created, which subsequently fragment into particle jets (left). The different particles within the jets are tracked by the LHCb detector. Selected features of the detected particle data are used as input for the Machine Learning analysis by NNs and TNs in order to determine the charge of the initial quark (right).

\paragraph*{Figure 2 --\textbf{Classification results}}
Comparison of the DNN and TNN analysis: (a) Tagging power for the DNN (green), TTN (blue) and the muon tagging (red), (b) ROC curves for the DNN (green) and the TTN (blue, but completely covered by DNN), compared with the \emph{line of no-discrimination} (dotted navy-blue line), (c) probability distribution for the DNN, and (d) for the TTN. In the two distributions (c)+(d), the correctly classified events (green) are shown in the total distribution (light blue). Below, in black all samples where a muon was detected in the jet.

\paragraph*{Figure 3 --\textbf{Exploiting the accessible information in a TTN}}
Exploiting the information provided by the learned TTN classifier: (a) Correlations between the $16$ input features (blue for anti-correlated, white for uncorrelated, red for correlated). The numbers indicate $q$, $p_T^{rel}$, $\Delta R$ of the muon (1-3), kaon (4-6), pion (7-9), electron (10-12), proton (13-15) and the jet charge $Q$ (16). 
(b) Entropy of each feature as the measure for the information provided for the classification. (c) Tagging power for learning on all features (blue), the best 8 proposed by QuIPS exploiting insights from (a)+(b) (magenta), the worst 8 (yellow) and the muon tagging (red). (d) Tagging power for decreasing bond dimension truncated after training: The complete model (blue shades for $\chi=100$, $\chi=50$, $\chi=5$), for using the QuIPS best 8 features only (violet shades for $\chi=16$, $\chi=5$), and the muon tagging (red).

\paragraph*{Figure 4 --\textbf{Tagging algorithms in LHCb}}

Illustrative sketch showing an LHCb experiment and the two possible tagging algorithms: a single particle tagging algorithm, exploiting information coming from one single particle (muon), and the inclusive tagging algorithm which exploits the information on all the jet constituents.

\paragraph*{Table 1 --\textbf{TTN prediction time}}
Prediction time, accuracy with (and without) applied cuts $\Delta$ and number of free parameters of the TTN for different bond dimension $\chi$ when we reduce the TTN model with QIANO, both for the complete 16 (left) and the QuIPS reduced 8 features (right). For the model $M_{16}$ with all 16 features (left), we trained the TTN with $\chi=200$ and truncate from there while for the reduced model $B_8$ (right), the original bond-dimension was $\chi=16$ (being the maximum $\chi$ in this subspace).

\end{document}